\documentclass{article}

\usepackage[utf8]{inputenc} 
\usepackage[T1]{fontenc}    
\usepackage{hyperref}       
\usepackage{url}            
\usepackage{booktabs}       
\usepackage{amsfonts}       
\usepackage{nicefrac}       
\usepackage{microtype}      
\usepackage{xcolor}         
\usepackage{graphicx}
\usepackage{algorithm}
\usepackage{algorithmic}
\usepackage{amsmath,amssymb,amsthm}
\usepackage{natbib}
\usepackage{verbatim}

\newcommand{\RR}{\mathbb R}

\title{Functional Labeled Optimal Partitioning}

\author{
  Toby D. Hocking, Toby.Hocking@nau.edu\footnote{Northern Arizona University}\\
  Jacob M. Kaufman, jmk478@nau.edu\footnotemark[\value{footnote}] \\ 
  Alyssa J. Stenberg, ajs937@nau.edu\footnotemark[\value{footnote}]
}

\begin{document}

\maketitle

\begin{abstract}
  Peak detection is a problem in sequential data analysis that involves differentiating regions with higher counts (peaks) from regions with lower counts (background noise). 
  It is crucial to correctly predict areas that deviate from the background noise, in both the train and test sets of labels. 
  Dynamic programming changepoint algorithms have been proposed to solve the peak detection problem by constraining the mean to alternatively increase and then decrease. 
  The current constrained changepoint algorithms only create predictions on the test set, while completely ignoring the train set. 
  Changepoint algorithms that are both accurate when fitting the train set, and make predictions on the test set, have been proposed but not in the context of peak detection models. 
  We propose to resolve these issues by creating a new dynamic programming algorithm, FLOPART, that has zero train label errors, and is able to provide highly accurate predictions on the test set. 
  We provide an empirical analysis that shows FLOPART has a similar time complexity while being more accurate than the existing algorithms in terms of train and test label errors.
\end{abstract}

\section{Introduction}

Changepoint detection models have become a prominent solution for fields such as genomics and medical monitoring that produce large amounts of data in a time or space series. 
Many of the proposed optimal changepoint models follow the classical dynamic programming algorithm of \citet{segment-neighborhood}, solving for the optimal $K$ segments ($K-1$ changes) in $n$ data in $O(Kn^2)$ time. 
This algorithm is optimal in the sense that it computes the changepoints and segment means which result in minimum loss given the data set.
Given a properly chosen non-negative penalty $\lambda$, the optimal solution can be computed in $O(n^2)$ time using the classic Optimal Partitioning algorithm of \citet{Jackson2005}.
Newer pruning methods have allowed the algorithm to decrease the number of changepoints assessed, which lowers the time complexity to $O(n\log n)$ while staying optimal \citep{Maidstone2016}. 
Binary segmentation is a classic heuristic algorithm which is often employed to reduce computation times, but it is not guaranteed to compute a set of changepoints with optimal loss \citep{binary-segmentation}.
In each of these algorithms there is a model complexity parameter (penalty or number of splits/segments), which can be chosen using either unsupervised \citep{Yao88, mBIC} as well as supervised methods when labeled training data are available \citep{Hocking2013icml, truong2017penalty}.
Labels are typically created by domain experts, and indicate presence or absence of changepoints in particular regions (Figure~\ref{fig:computation-graph}).

\subsection{Algorithms for peak detection}
This paper, however, focuses on changepoint models with constrained model parameters to produce more interpretable results. 
One problem that motivated the introduction of constrained models is peak detection in ChIP-seq data \citep{Barski2007}. 
Peak detection is the problem of differentiating regions with higher counts (peaks) from areas with lower counts (background noise). 
Up-down constrained changepoint models can also be computed in $O(n\log n)$ time, and have shown high peak detection accuracies in ChIP-seq data \citep{Hocking2020jmlr, RungeGFPOP}. 
The up-down constraint on the model forces an up change in the segment mean parameter to be followed by a down transition and vice versa. 
This constraint is necessary for interpretation as a peak detection model because, in an unconstrained changepoint model, the outputted segmentation can result in consecutive up-down changes in the segment mean parameter.
In recent findings, the max jump post-processing rule has improved upon the accuracy of the up-down constrained models in ChIP-seq data \citep{max-jump-rule}. The max jump rule converts the outputted segmentation of unconstrained changepoint models into peaks and background states. The post-processing rule works by grouping consecutive changes in segment mean parameters (either up or down) together and choosing the maximum mean difference from each group.

\begin{table}
  \caption{Comparison between proposed FLOPART algorithm and previous  algorithms for changepoint detection (column for publication, label constraints, up-down constraint, and time). }
  \label{novelty-table}
  \centering
  \begin{tabular}{lllll}
    \toprule
    \cmidrule(r){1-2}
    Algorithm & Publication & Labels & Up-down & Time \\
    \midrule
    OPART  & \citet{Jackson2005}  & No & No & $O(n^2)$   \\
    LOPART  & \citet{Hocking2020LOPART}  & Yes & No & $O(n^2)$   \\
    SegAnnot  & \citet{Hocking2012}  & Yes & No & $O(Kn^2)$   \\
    BINSEG  & \citet{binary-segmentation}  & No & No & $O(n\log K)$  \\
    FPOP  & \citet{Maidstone2016}  & No & No & $O(n\log n)$   \\
    GFPOP  & \citet{RungeGFPOP}  & No & Yes & $O(n\log n)$   \\
    FLOPART  & {\bf This paper}  & Yes & Yes & $O(n\log n)$   \\
    \bottomrule
  \end{tabular}
\end{table}

\paragraph{Novelty with respect to previous work.}
We propose a new supervised up-down constrained changepoint detection algorithm for the case of labeled data sequences.
The current fastest constrained algorithm, GFPOP  \citep{RungeGFPOP}, can be considered unsupervised because the algorithm does not use the labels in the train set, allowing for train errors. 
In contrast, changepoint models that currently utilize labels are all unconstrained models (no constraints between adjacent segment means), including the LOPART algorithm \citep{Hocking2020LOPART} and the SegAnnot algorithm \citep{Hocking2012}. 
In this paper, we resolve the drawbacks of both previous algorithms, resulting in our new Functional Labeled Optimal PARTitioning (FLOPART) algorithm, which incorporates the label constraint ideas from LOPART with the constraints on adjacent segment means from GFPOP (Table~\ref{novelty-table}).

\section{Mathematical framework and optimization problems}
\subsection{Previous problem with label constraints but no constraints on adjacent segment means}
We begin this section by explaining the optimization problem solved by the previous LOPART algorithm \citep{Hocking2020LOPART}, which has label constraints but no up-down constraints.
We are given a data sequence $z_1,\dots,z_n$, and 
$l$ labeled regions
$(\underline p_1, \overline p_1, t_1), \dots,
(\underline p_l, \overline p_l, t_l)$, where
\begin{itemize}
\item $\underline p_j<\overline p_j$ are the start/end of a labeled region (in data sequence coordinates, 1 to $n$),
\item $t_j\in\{0,1\}$ is the type of change, here the number of changes in the region
  $[\underline p_j, \overline p_j]$. 
\end{itemize} 
We assume the labels are ordered, $1\leq \underline p_1< \overline p_1 \leq \cdots \leq \underline p_l < \overline p_l \leq n$.
We would like to compute a mean vector $\mathbf m\in\mathbb R^n$ which solves the following optimization problem,
\begin{align}
\min_{
  \mathbf m\in\RR^{n}
  } &\ \ 
\sum_{i=1}^n \ell(z_i, m_i) +
\lambda\sum_{i=1}^{n-1} I(m_i \neq m_{i+1})
  \label{LOPART-objective}
\\
    \text{subject to} 
& \ \ \sum_{i=\underline p_j}^{\overline p_j-1} I(m_i \neq m_{i+1})=t_j
\text{ for all } j\in\{1,\dots,l\},
\label{LOPART-label-constraints}
\end{align}
where $\ell$ is a loss function and $I$ is the indicator function (1 if true, 0 otherwise).
The objective~(\ref{LOPART-objective}) in the optimization problem above contains terms for the 
loss $\ell$ (to encourage data fitting), and a per-change penalty of
$\lambda\geq 0$ (to control model complexity). 
There are also constraints~(\ref{LOPART-label-constraints}) which ensure that the correct number of changes occur in each label.

SegAnnot is one algorithm that can be used in this context
 \citep{Hocking2012}, but it is
limited in that it can not detect any changes that are not labeled, which corresponds to using a large value for the penalty $\lambda$.
The LOPART algorithm can compute a globally optimal solution to this problem, for any $\lambda$, in $O(n^2)$ time \citep{Hocking2020LOPART}.

\begin{figure}
    \centering
    \includegraphics[width=\textwidth]{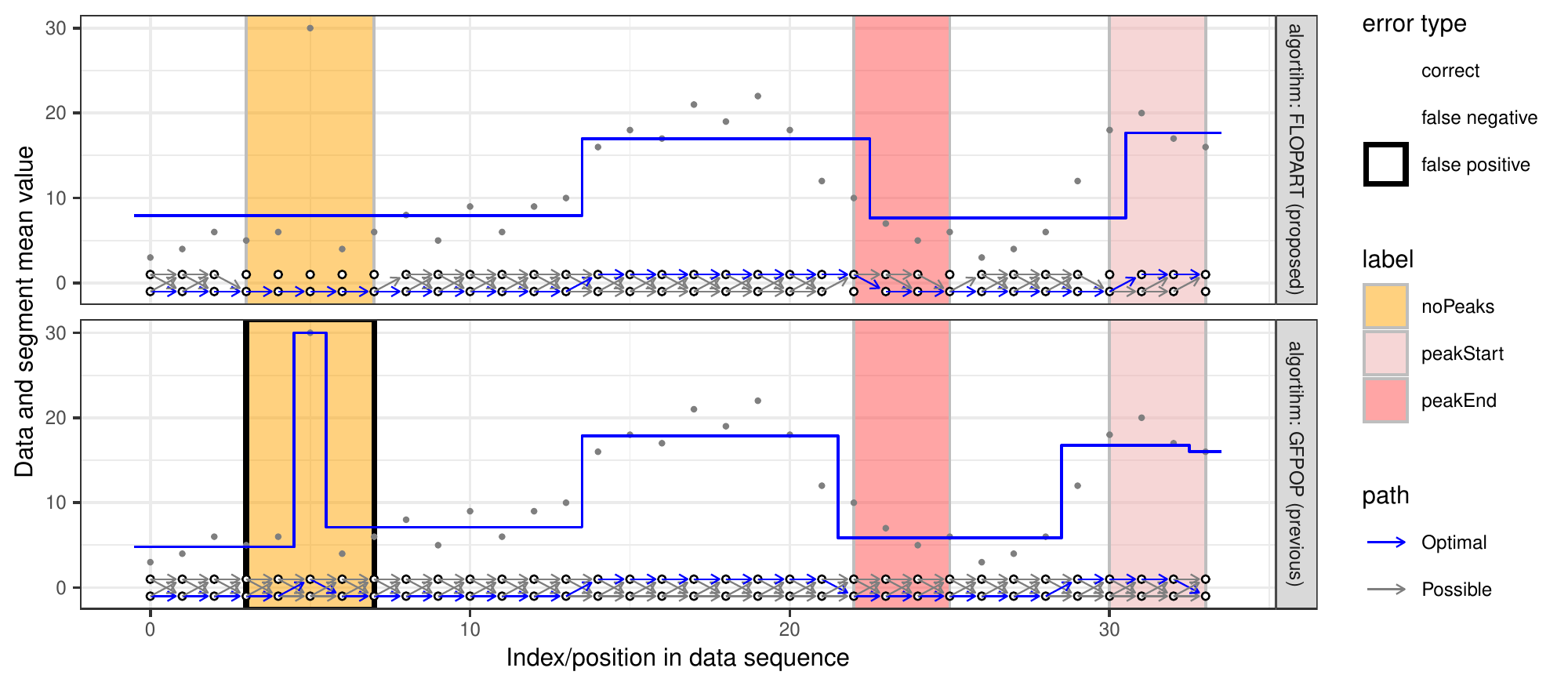}
    \vspace{-0.7cm}
    \caption{Data (grey dots), segmentation (blue lines), labels (colored rectangles), and computation graph (bottom of each panel, one node for each cost function, top row for peak state, bottom row for background state).
    Whereas the previous GFPOP algorithm (bottom) does not use the labels so has three label errors, the proposed FLOPART algorithm (top) constrains the computation graph so that the segmentation is consistent with the labels.
    }
    \label{fig:computation-graph}
\end{figure}

\subsection{Proposed problem with label constraints and alternating up/down constraints}

In the context of weakly supervised peak detection in a sequence of $n$ data \citep{Hocking2020jmlr}, the goal is to predict a sequence of $n$ binary values (0 for no peak, 1 for peak). 
We have three types of labels: noPeaks $(t_j=0)$ indicates there should be no peaks in $[\underline p_j,\overline p_j]$, peakStart $(t_j=1)$ indicates that there should be one change up in $[\underline p_j,\overline p_j]$, and peakEnd $(t_j=-1)$ indicates that there should be one change down in $[\underline p_j,\overline p_j]$.
The optimization variables are the segment means $\mathbf m=[m_1, \dots, m_N]$, hidden states $\mathbf s=[s_1, \dots, s_N]$, and changepoints $\mathbf c=[c_1, \dots, c_{N-1}]$,
\begin{align}
  \label{min:op-up-down}
  \min_{
  \substack{
  \mathbf m\in\RR^N,\ \mathbf s\in\{0, 1\}^N\\
  \mathbf c\in\{-1, 0,1\}^{N-1}\\
  }
  } &\ \
      \sum_{i=1}^N \ell( z_i, m_i) + \lambda \sum_{i=1}^{N-1} I(c_i \neq 1) \\
  \text{subject to\ \ } &\ \text{no change: }c_i = 0 \Rightarrow m_i = m_{i+1}\text{ and }s_i=s_{i+1}, \nonumber\\
    &\ \text{non-decreasing change: }c_i = 1 \Rightarrow m_i \leq m_{i+1}\text{ and }(s_i,s_{i+1})=(0,1),\nonumber\\
    &\ \text{non-increasing change: } c_i = -1 \Rightarrow m_i \geq m_{i+1}\text{ and }(s_i,s_{i+1})=(1,0),\nonumber\\
  & \ \text{peakStart/peakEnd label constraints: } 
  \forall t_j\in\{-1,1\},\, 
  \sum_{i=\underline p_j}^{\overline p_j} I[c_i = t_j] = 1,\label{peak-start-end-constraints}\\
  & \ \text{noPeaks label constraints: }
  \forall t_j=0,\, \forall i\in [\underline p_j,\overline p_j],\,
  s_i=0.
  \label{noPeaks-label-constraints}
\end{align}
The optimization problem~(\ref{min:op-up-down}) is essentially the same as for the GFPOP algorithm \citep{RungeGFPOP} but with added constraints (\ref{peak-start-end-constraints}--\ref{noPeaks-label-constraints}) which ensure that the model is consistent with the labels.
Constraint~(\ref{peak-start-end-constraints}) ensures that there is one up change in each peakStart label, and one down change in each peakEnd label.
Constraint~(\ref{noPeaks-label-constraints} ensures that the background state ($s_i=0$) is always predicted in each noPeaks label.

\section{New dynamic programming algorithm}

In this section we propose a new algorithm, FLOPART, which computes a globally optimal solution to a simplified version of problem~(\ref{min:op-up-down}). 
The simplification that we propose is additional constraints that ensure the model starts down and ends up for each peakStart label, and the model starts up and ends down for each peakEnd label.
This results in a new algorithm that ensures the label constraints are respected in problem~(\ref{min:op-up-down}), and which is easy to implement and interpret in terms of removing edges in the computation graph (Figure~\ref{fig:computation-graph}).

\subsection{Dynamic programming update rules}
The GFPOP dynamic programming algorithm for changepoint detection requires computation of the optimal cost as a function of the last segment mean, which can be exactly represented for common loss functions like the square loss and Poisson loss \citep{RungeGFPOP}.
In the case of our up-down constrained peak detection model there are two states:
up/peak ($s_i=1$) and down/background ($s_i=0$). 
We therefore need to compute two kinds of cost functions, $C_{0,i}(\mu),C_{1,i}(\mu)$ for all
$i\in\{1,\dots,n\}$. 
Each $C_{s,i}(\mu)$ function is the optimal cost up to data point $i\in\{1,\dots,n\}$, in state $s\in\{0,1\}$, with last segment mean $\mu\in\mathbb R$.
These cost functions can be recursively computed using update rules which can be visualized as nodes and edges in a computation graph (Figure~\ref{fig:computation-graph}).
There is a node in the computation graph for each cost function $C_{s,i}$.
There are one or more edges directed toward each node; each of these edges represents a previously computed cost function which is used to compute the cost at that node.
The dynamic programming algorithm can be interpreted as finding the path with minimum cost in this computation graph.

\paragraph{Initialization and previous update rule for unlabeled regions.} The first cost functions are initialized to the un-penalized loss of the first data point, $C_{0,0}(\mu)=C_{1,0}(\mu)=\ell(z_0,\mu)$.
For any $i>0$ in an unlabeled region, the update rules are the same as for GFPOP (with up-down constraints but no label constraints),
\begin{eqnarray}
  C_{0,i}(\mu)
  &=& \ell(z_i, \mu) + \min\{
       C_{0,i-1}(\mu),
       C_{1,i-1}^\geq(\mu) +\lambda
      \}, \label{C_0i_nolabel}\\
        C_{1,i}(\mu)
  &=& \ell(z_i, \mu) + \min\{
       C_{1,i-1}(\mu),
       C_{0,i-1}^\leq(\mu) + \lambda
      \}, \label{C_1i_nolabel}
\end{eqnarray}
where $C_{0,i-1}^\leq(\mu)=\min_{x\leq \mu} C_{0,i-1}(x)$ in (\ref{C_1i_nolabel}) is the ``min-less operator,'' which is used to compute the optimal cost of a change up, as previously described \citep{Hocking2020jmlr}. 
Similarly, $C_{1,i-1}^\geq(\mu)=\min_{x\geq \mu} C_{1,i-1}(x)$ in (\ref{C_0i_nolabel}) is the ``min-more operator,'' which is used to compute the optimal cost of a change down.
The update rules~(\ref{C_0i_nolabel}--\ref{C_1i_nolabel}) can be used to compute the cost functions at data point $i$, given the cost functions at the previous data point $i-1$.
These update rules can be seen in Figure~\ref{fig:computation-graph} --- for all $i$ with GFPOP (bottom), and for all $i$ outside of labels for FLOPART (top), each node has edges coming from the node(s) with finite cost at the previous data point.

\paragraph{Proposed update rules for labeled regions.} 
For $i\in[\underline p_j,\overline p_j]$ in a $t_j=0$ (noPeaks) label, we have peak state cost $C_{1,i}(\mu) = \infty$ (it is not possible to be in the up/peak state), and the update rule for the background state cost $C_{0,i}$ is the same as in the unlabeled case~(\ref{C_0i_nolabel}). 
Because the cost of being in the up/peak state is infinite, the computation graph can be simplified by removing edges. 
This update rule can be seen in Figure~\ref{fig:computation-graph} (top), for all $i$ in the noPeaks label, there are no edges connected to peak state nodes.

For $i\in[\underline p_j,\overline p_j]$ in a $t_j=1$ (peakStart) label, we simplify problem~(\ref{min:op-up-down}) by assuming we need to be in the down state at the start of the label and the up state at the end of the label.
The corresponding update rules can be seen in Figure~\ref{fig:computation-graph} (top); in the peakStart label, we have (i) the first up cost node has no connected edges, (ii) there are no edges from up to down cost nodes, and (iii) the last down cost node has no connected edges.
The update rules are therefore
\begin{eqnarray}
   C_{0,i}(\mu)
  &=& \ell(z_i, \mu) + \begin{cases}
    \min\{
     C_{0,i-1}(\mu),
     C_{1,i-1}^\geq(\mu) +\lambda
    \} & \text{ if } i=\underline p_j,\\
    \infty & \text{ if }i=\overline p_j,\\
     C_{0,i-1}(\mu) & \text{ otherwise.}\label{C_0i_peakStart}
  \end{cases} \\
  \label{eq:c1peak}
   C_{1,i}(\mu)
  &=& \ell(z_i, \mu) +\begin{cases}
    \infty & \text{ if } t=\underline p_j,\\
     \min\{
     C_{1,i-1}(\mu),
     C_{0,i-1}^\leq(\mu) +\lambda
    \} &  \text{ otherwise.}\label{C_1i_peakStart}
  \end{cases} 
\end{eqnarray}
Update rule~(\ref{C_0i_peakStart}) for computing the down cost $C_{0,i}(\mu)$ in a peakStart label has three cases.
At the start of the peakStart label ($i=\underline p_j$), the same unlabeled update rule~(\ref{C_0i_nolabel}) is used to compute the down cost (either no change from a previous down cost, or a change down from a previous up cost).
At the end of the label ($i=\overline p_j$), the down cost is infinite (only the up cost is feasible at the end of the peakStart label).
In other positions $i$ inside the peakStart label, only the previous down cost is used (no change down from a previous up cost allowed).
Update rule~(\ref{C_1i_peakStart}) for computing the up cost $C_{1,i}(\mu)$ in a peakStart label has two cases.
At the start of the peakStart label ($i=\underline p_j$), the up cost is infinite (only the down cost is feasible at the start of a peakStart label).
For other positions $i$ inside the peakStart label, the same unlabeled update rule~(\ref{C_1i_nolabel}) is used to compute the up cost (either no change from a previous up cost, or a change up from a previous down cost).
The update rule for positions $i$ in a $t_j=-1$ (peakEnd) label is analogous.

Note that given the update rules above, labels could be inconsistent if $\overline p_j = \underline p_{j+1}$.
For example, two peakStart labels right next to each other would be inconsistent since each must start down and end up (all cost functions after the last $i$ in the first peakStart label would be infinite). 
To deal with that, we can enforce $\overline p_j < \underline p_{j+1}$ in the label positions, and check that before running the dynamic programming algorithm.

\subsection{Explanation of pseudo-code}
On line \ref{line:data_loop}, FLOPART loops over each data point to fill in the cost matrix. Inside the data loop, on line \ref{line:labelNum}, the GetLabel subroutine determines which label type applies to the current data point $i$ --- peakStart, peakEnd, noPeaks, or unlabeled. 
Then, on line \ref{line:state_loop}, the algorithm loops over the possible states of the current data point --- background or peak. Inside the state loop, on line \ref{line:getLabel} the algorithm will determine the rule to calculate the cost of being in state $s$ at the current data point $i$ based on the label type determined on line \ref{line:labelNum}. 

Then, on line \ref{line:getCost}, the $C_{s,i}$ entry in the cost matrix is filled in based on the rule determined on line \ref{line:getLabel}, the cost functions $C_{0,i-1}, C_{1,i-1}$ at the previous data point, the current data value $z_i$, and the per-change penalty $\lambda$. 
The equations used to determine the cost functions are (\ref{C_0i_nolabel}--\ref{C_1i_nolabel}) if point $i$ is unlabeled,  (\ref{C_0i_peakStart}--\ref{C_1i_peakStart}) for a peakStart label, etc. 
After filling in the cost matrix with the costs of being in states 0 or 1 for each data point based on the label constraints $L$, then the optimal segment means, segment ends, and states of each data point are determined by the Decode sub-routine (line~\ref{line:output}), which is implemented as previously described \citep{RungeGFPOP}.

\begin{algorithm}[t]
\begin{algorithmic}[1]
\STATE Input: data $\mathbf z\in\mathbb R^n$, 
penalty $\lambda\in\RR_+$,
sorted labels $L = (\underline p_1, \overline p_1, t_1),\dots,(\underline p_l, \overline p_l, t_l)$.
\STATE Allocate $2\times n$ matrix of cost functions $C = [C_{0,1},\dots,C_{0,n}|C_{1,1},\dots,C_{1,n}]$.
\STATE for data index $i$ from $1$ to $n$: \label{line:data_loop} 
\begin{ALC@g}
  \STATE $\text{labelNum}\gets\text{GetLabel}(i,L)$ \label{line:labelNum}
  \STATE for state $s$ from $0$ to $1$: \label{line:state_loop} 
  \begin{ALC@g}
    \STATE $\text{rule}\gets \text{GetRule}(s,i,L,\text{labelNum})$ \label{line:getLabel}
    \STATE $C_{s,i}\gets  \text{rule.GetCost}(C_{0,i-1}, C_{1,i-1},z_i,\lambda)$  \label{line:getCost}
  \end{ALC@g}
\end{ALC@g}
\STATE Output: $\text{means},\text{segment\_ends},\text{states}\gets\text{Decode}(C)$ \label{line:output}
\caption{\label{algo:LabeledFPOP}Functional Labeled Pruning Optimal
  Partitioning Algorithm (FLOPART).}
\end{algorithmic}
\end{algorithm}

\section{Empirical Results}
\subsection{Data set and baseline algorithms}
\label{sec:datasets}
To examine the changepoint prediction accuracy of FLOPART, we performed the following experiments using real ChIP-seq peak detection data sets. For baseline algorithms, we considered GFPOP \citep{RungeGFPOP}, BINSEG \citep{binary-segmentation}, and LOPART \citep{Hocking2020LOPART}.  
Expert biologists created labels for seven labeled ChIP-seq data sets using visual inspection to determine if there were/weren't significant peaks in particular genomic regions \citep{Hocking2016ChIP}. 
The data consists of seven labeled ChIP-seq split by two peak types, H3K4me3 (sharp peak pattern) and H3K36me3 (broad peak pattern); each peak is of interest as they indicate genomic regions with active genes \citep{histone-review}. 
Each peak data set includes multiple samples of different cell types, including tcell, bcell, and monocyte. 
We ran all the experiments using a MacBook Pro with the 8-core CPU M1 chip and 16GB of unified memory. 

\begin{figure}
    \centering
    \includegraphics[width=\textwidth]{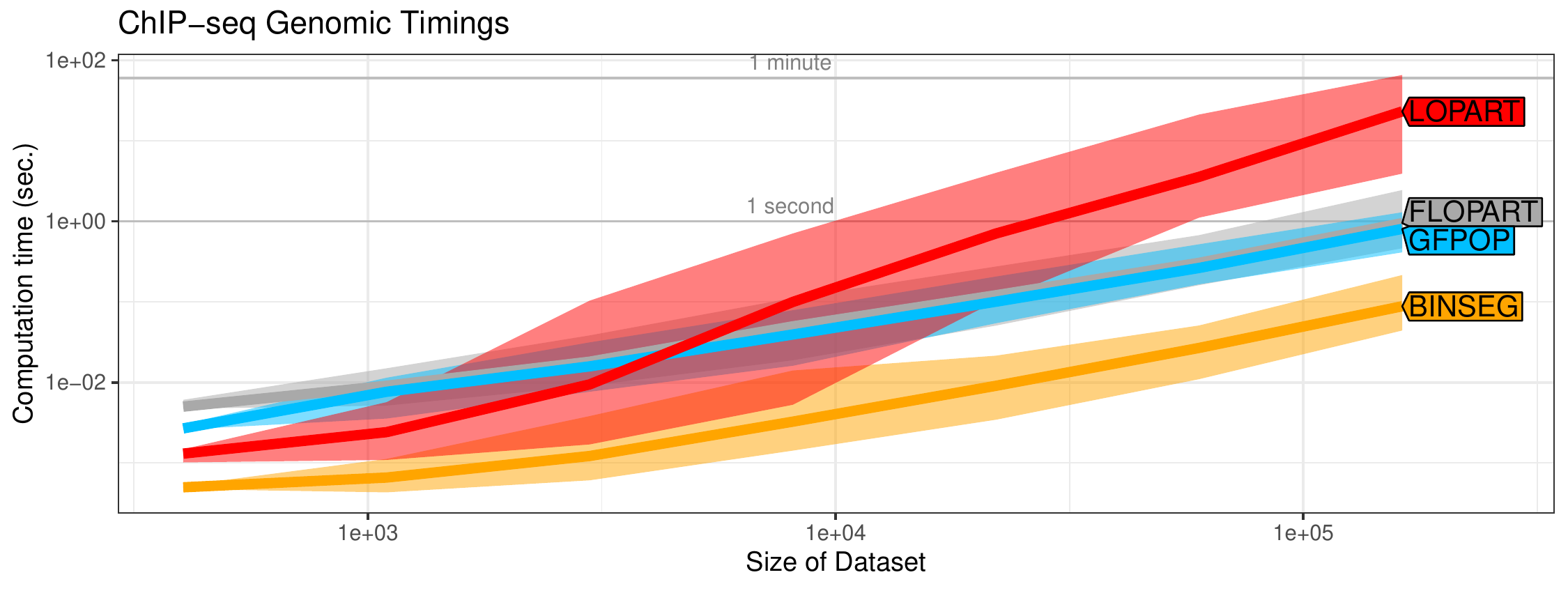}
    \vspace{-0.7cm}
    \caption{Empirical timings in real ChIP-seq data sets with FLOPART and the baseline algorithms (median line and max/min band computed over all the datasets from section \ref{sec:datasets}).
    FLOPART shows log-linear time complexity, much faster than the previous quadratic time LOPART algorithm.
    }
    \label{fig:timings}
\end{figure}

\subsection{Empirical time complexity in real data sets}

FLOPART uses a similar algorithm as GFPOP (dynamic programming using functional pruning), so we suspected the two algorithms should have the same log-linear time complexity. 
To verify this empirically, we conducted a timings experiment using the real ChIP-seq datasets and baseline algorithms from section \ref{sec:datasets}. 
We timed how long each algorithm took to run on each data sequence and then plotted the time each algorithm took to run against the size of each data sequence (Figure \ref{fig:timings}). 
From this data, we saw FLOPART runs asymptotically faster than LOPART, which is quadratic $O(n^2)$ time. 
The data also shows that FLOPART had the same asymptotic slope as GFPOP and BINSEG (only different by small constant factors), which are log-linear $O(n\log n)$ time algorithms. 
Overall this experiment shows that FLOPART is as fast as GFPOP and BINSEG (up to asymptotic constant factors), and is much faster than LOPART.

\subsection{Analysis of best case label error in real genomic data}

\begin{figure}
    \centerline{
        \includegraphics[height = 2in]{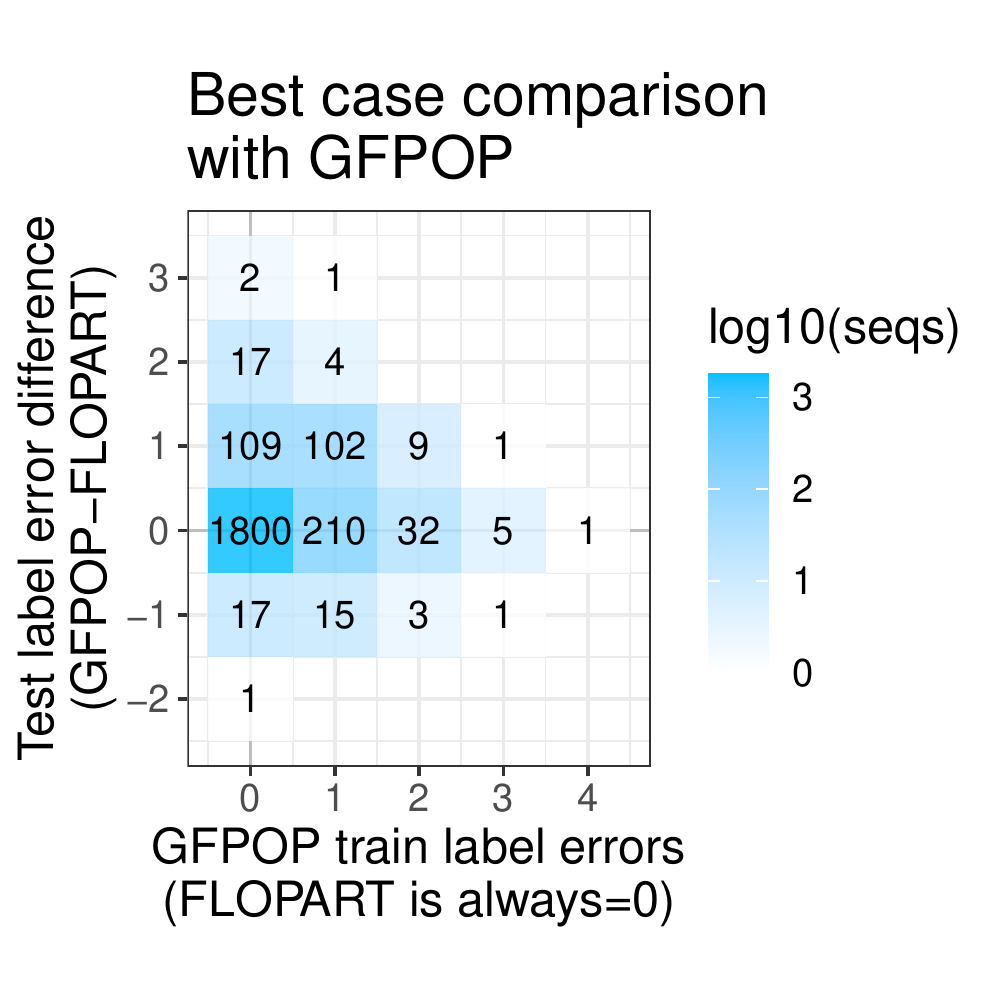}
        \includegraphics[height = 2in]{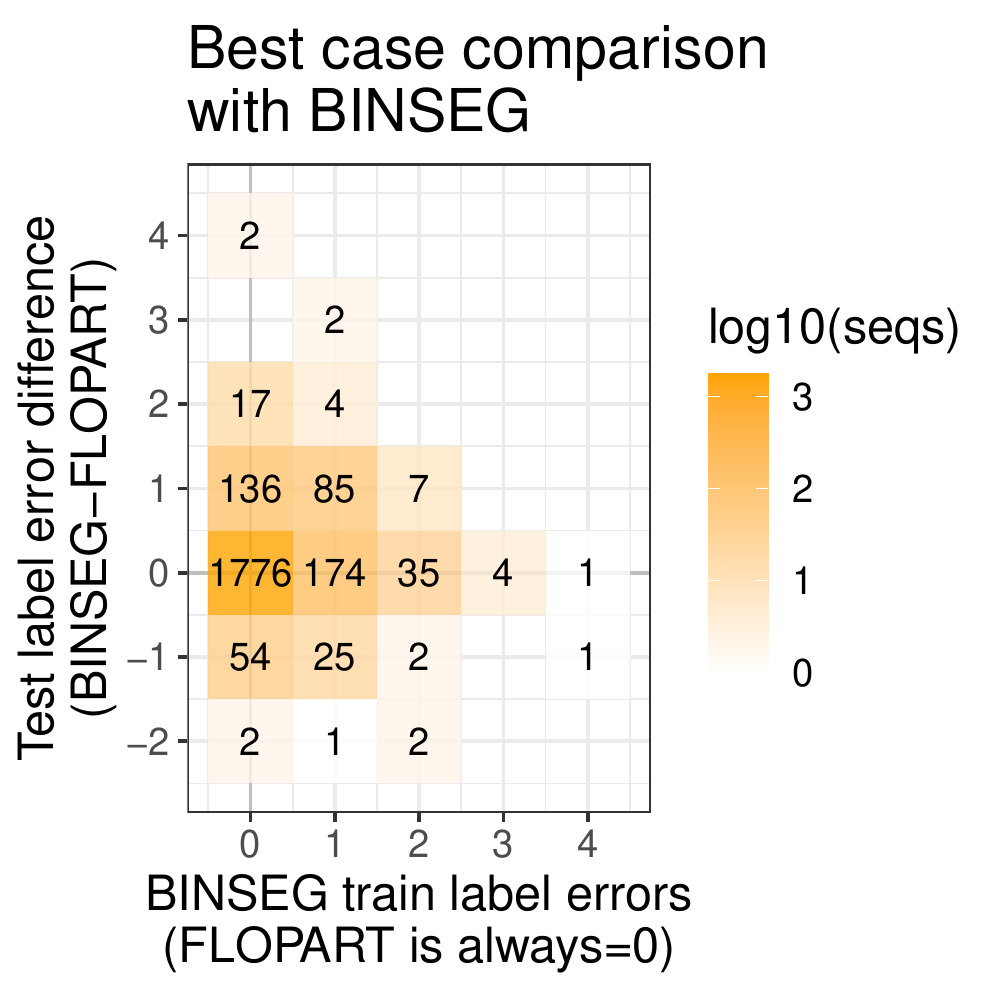}
        \includegraphics[height = 2in]{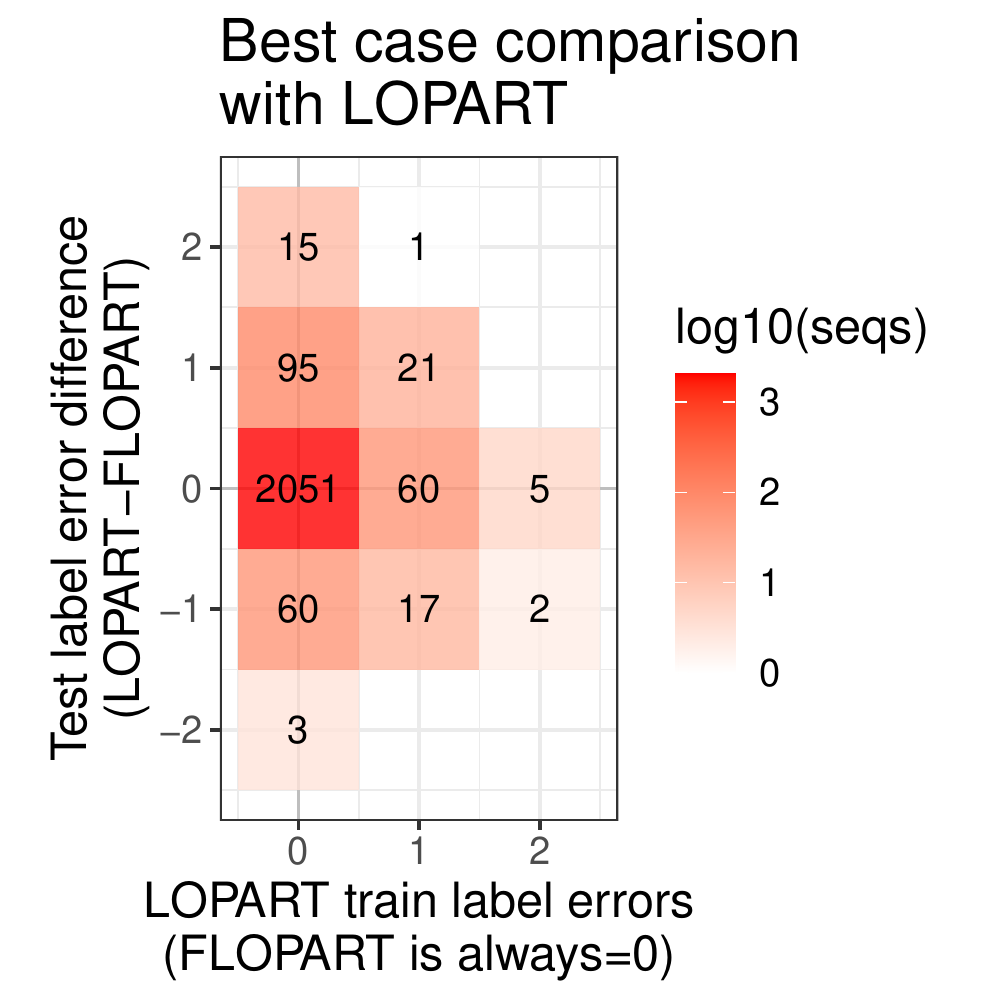}
    }
    \caption{Comparing FLOPART to baseline algorithms in terms of number of label errors in the best case in real ChIP-seq data sets (selected the $\lambda$ penalty over each algorithm/sequence/split by minimizing total label errors (train+test).
    }
    \label{fig:label-errors}
\end{figure}

For this analysis, we looked at the label error difference between FLOPART and the baseline algorithms in the best case for each algorithm.
To start, we randomly assigned each label in every ChIP-seq dataset to a random fold ID and used $K = 2$ cross-validation to get two train/test splits per sequence (each train/test split contained at least one label). 
On each of the 2,330 labeled ChIP-seq sets and train/test split, we ran FLOPART and the baseline algorithms on a grid of 23 penalty values evenly spaced on the log scale $\lambda\in\{10^{-5}, 10^{-4.5}, \dots, 10^6\}$. 
For every split/sequence/algorithm, we chose the penalty value that minimized the total number of label errors (train+test), and examined the difference in number of label errors between FLOPART and the baseline algorithms (Figure~\ref{fig:label-errors}). 
We also tried first minimizing train errors then minimizing test errors, and results were qualitatively similar to the results we report below by minimizing total label errors (train+test).

\paragraph{Best penalty comparison with GFPOP}
Since FLOPART has zero errors on the train set by definition, we predicted GFPOP to have more errors in some data sequences. We observed that with the best-predicted penalty; GFPOP had 0 training errors in $ 1,946/2,329 = 84\% $ of datasets/sample ID/folds (counts on the vertical line in Figure \ref{fig:label-errors}, left) but had $1-4$ train set errors $384/2,329 = 16\% $ of sequences (counts on the right vertical line in Figure \ref{fig:label-errors}, left). Looking at the test set errors per algorithm, we observed FLOPART had the same number of test set errors in $2,048/2,329 = 88\% $ datasets/sample ID/folds (counts on the horizontal line in Figure \ref{fig:label-errors}, left). We also observed that FLOPART had $1-3$ fewer test set errors in $245/2,329 = 11\% $ sequences (counts above the horizontal line in Figure \ref{fig:label-errors}, left) while only having $1-2$ more test errors in $37/2,329 = 1\% $ sequences (counts below the horizontal line in Figure \ref{fig:label-errors}, left). This data is a clear indication that after picking the best case penalty for each model (FLOPART and GFPOP), FLOPART $99\% $ of the time will be as accurate as GFPOP in these real genomic datasets, and sometimes more accurate. 

\paragraph{Best penalty comparison with BINSEG}
Since FLOPART has zero errors on the train set by definition, and BINSEG is not guaranteed to solve the changepoint optimization problem, we expected BINSEG should have more errors in some data sequences. 
We observed that with the best-predicted penalty, BINSEG had 0 training errors in $ 1,987/2,329 = 85\% $ of datasets/sample ID/folds (counts on the vertical line in Figure \ref{fig:label-errors}, middle) but had $1-4$ train set errors $343/2,329 = 15\% $ of sequences (counts right of vertical line in Figure \ref{fig:label-errors}, middle). Looking at the test set errors per algorithm, we observed FLOPART had the same number of test set errors in $1,993/2,329 = 86\% $ datasets/sample ID/folds (counts on the horizontal line in Figure \ref{fig:label-errors}, middle). We also observed that FLOPART had $1-4$ fewer test set errors in $253/2,329 = 11\% $ sequences (counts above the horizontal line in Figure \ref{fig:label-errors}, middle) while only having $1-2$ more test errors in $87/2,329 = 3\% $ sequences (counts below the horizontal line in Figure \ref{fig:label-errors}, middle). Looking at the data we see the FLOPART is as accurate as BINSEG $97\%$ of the time and, in some cases more accurate than BINSEG by $1-4$ test set errors.

\paragraph{Best penalty comparison with LOPART}
Since FLOPART and LOPART are supposed to have zero errors on the train set by definition, we predicted that they would be comparable with train and test set errors in most data sequences. We observed that with the best-predicted penalty; LOPART had 0 training errors in $ 2,224/2,329 = 95\% $ of datasets/sample ID/folds (counts on the vertical line in Figure \ref{fig:label-errors}, right) but had $1-2$ train set errors $106/2,329 = 5\% $ of sequences (counts on the right vertical line in Figure \ref{fig:label-errors}, right). We found that using LOPART with the max jump rule does not guarantee $100\%$ accuracy on the train set. Translating the outputted segments into peaks and background states sometimes causes the labels to be incorrect due to combining consecutive jumps and predicting the wrong state because of the alternating states of peak detection models. Looking at the test set errors per algorithm, we observed FLOPART had the same number of test set errors in $2,116/2,329 = 91\% $ datasets/sample ID/folds (counts on the horizontal line in Figure \ref{fig:label-errors}, right). We also observed that FLOPART had $1-2$ fewer test set errors in $132/2,329 = 6\% $ sequences (counts above the horizontal line in Figure \ref{fig:label-errors}, right) while only having $1-2$ more test errors in $82/2,329 = 3\% $ sequences (counts below the horizontal line in Figure \ref{fig:label-errors}, right). Looking at the data we see the FLOPART is as accurate as LOPART $97\%$ of the time and, in some cases more accurate than LOPART by $1-2$ test set errors. 

\subsection{Analysis of test label errors in real genomic data when learning/predicting the penalty}

\begin{figure}
    \centering
    \includegraphics[width=\textwidth]{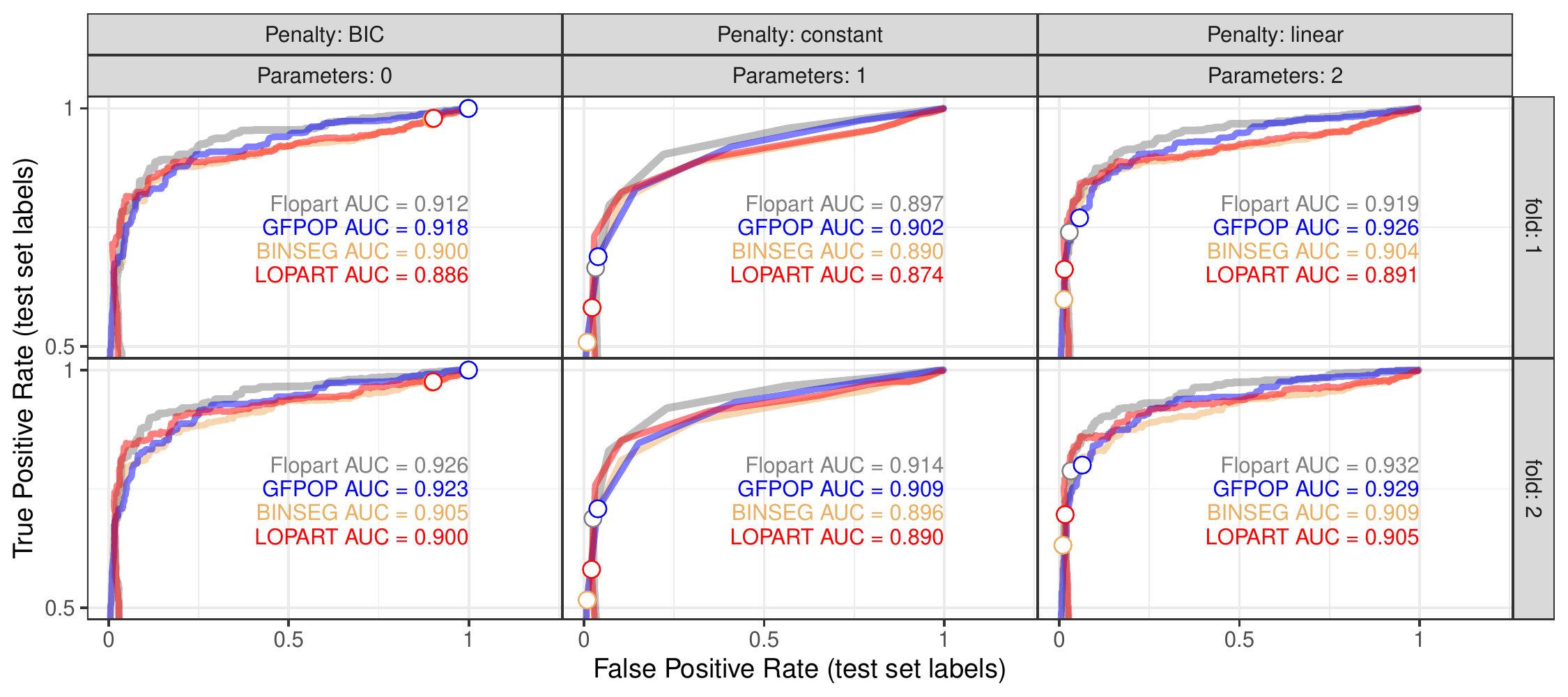}
    \vspace{-0.7cm}
    \caption{Receiver Operating Characteristic (ROC) analysis of penalty predictions using three penalty learning methods (left to right) and two cross-validation folds (top to bottom) on real ChIP-seq data. 
    Default predictions  $f(x_i)=\log\lambda_i$ are shown as white dots, and different points on each ROC curve are obtained by varying a constant $c$ added to predicted log penalty values, $f(x_i)+c$.
    FLOPART always has a larger Area Under the Curve (AUC) than BINSEG and LOPART. FLOPART also has a comparable AUC to GFPOP over the two folds.}
    \label{fig:Roc}
\end{figure}

In this section we study the accuracy of FLOPART when using a penalty value learned using GFPOP (which is the same as running FLOPART with no labels). 
To choose the penalty value we used with FLOPART, we ran GFPOP using the same penalty grid as the best penalty analysis, then computed label error rates for each dataset/sample ID/penalty/split (good penalties are those with small label error rates on the train data). 
For comparison, we did the same for BINSEG and LOPART (with no labels input to LOPART).
We used three different methods for predicting penalty values $\lambda$ to run with each dataset and sample ID, as previously described \citep{Hocking2020LOPART}. 
Each method can be viewed as learning a function $f(x_i)=\log\lambda_i$ which predicts a log penalty value using features $x_i$ for test example $i$.
\begin{description}
\item[BIC] uses the Bayesian Information Criterion of \citet{Schwarz78} by choosing a $\lambda_i=\log N_i$ for each data set $i$ where $N_i$ is the total number of weights to segment --- $f(x_i)=\log\lambda_i=\log\log N_i$, zero learned parameters (unsupervised).
\item[constant] searches the grid of penalties for the minimum train label errors then selects that penalty value as a constant penalty, $\lambda$ for each test data set --- $f(x_i)=\log\lambda$, one learned parameter.
\item[linear] uses a single feature $x_i = \log \log N_i$ for each data set $i$, then optimizes a squared hinge loss using gradient descent \citet{Hocking2013icml} to learn weight $w$ and bias $b$ parameters --- $ f(x_i) = \log \lambda_i  = w^T x_i + b$, two learned parameters.
\end{description}
We ran GFPOP and FLOPART on each test data sequence $i$, using each of the three different penalty prediction methods $f(x_i)=\log \lambda_i$ (BIC/constant/linear).
To perform Receiver Operating Characteristic (ROC) curve analysis (Figure~\ref{fig:Roc}), we vary a constant $c$ added to the predicted log penalty, $f(x_i)+c$, to obtain the different points on the ROC curve, and then compare accuracy of different methods using the Area Under the ROC curve (AUC, larger values for more accurate methods).

\paragraph{Predicted penalty comparison with GFPOP and LOPART}
We expected that using a learned penalty value from GFPOP with FLOPART should result in comparable prediction accuracies with GFPOP and LOPART because all three algorithms utilize penalty values identically. 
Looking at the ROC curve analysis (Figure \ref{fig:Roc}), we found that FLOPART was more accurate than LOPART in both folds. FLOPART had a $0.023- 0.029$ larger AUC value than LOPART, which makes sense because FLOPART has an informative prior (up-down constraints) whereas LOPART does not. 
FLOPART compared more closely to GFPOP, only having a $0.003-0.005$ larger AUC value for fold two, and having a $0.005-0.007$ smaller AUC value for fold one. 
This data shows that a penalty learned from GFPOP can be used for predictions using FLOPART. 
It also indicates that FLOPART has more accurate predictions than LOPART and compares well to GFPOP. 

\paragraph{Predicted penalty comparison with BINSEG}
Since BINSEG is a heuristic algorithm which is not guaranteed to compute optimal changepoints, we predicted that FLOPART should have more accurate predictions than BINSEG. 
Looking at the ROC curve analysis (Figure \ref{fig:Roc}), we see that FLOPART had $0.007-0.023$ larger AUC than BINSEG. 
These data show that FLOPART has slightly more accurate predictions than BINSEG.

\section{Discussion and Conclusions}
We present a new algorithm, FLOPART, for peak detection in labeled time series data. 
It combines ideas from Generalized Functional Optimal Partitioning (GFPOP) \citet{RungeGFPOP} with Labeled Optimal Partitioning (LOPART) \citet{Hocking2020LOPART}. The previously proposed labeled changepoint detection model LOPART guarantees accuracy with train set labels but runs in quadratic time and does not have the up-down constraint needed for compatibility with peak detection models. The novelty of FLOPART concerning LOPART is the up-down constraint on the segment mean parameter, which allows FLOPART to be accurate and interpretable as a peak detection model.

Our empirical timings analysis using ChIP-seq data showed that FLOPART runs much faster than the previous algorithm LOPART and runs in log-linear $O(n\log n)$ time, like the unlabeled competitors GFPOP and BINSEG. 
Our empirical accuracy analysis using best case penalties in ChIP-seq data showed that for a train set, FLOPART is always at least as accurate as of the GFPOP, LOPART, and BINSEG baselines and frequently even more accurate. 
Finally, the predicted penalty experiment displayed that FLOPART is still accurate using a penalty learned from GFPOP for predictions.
We even noticed that FLOPART is more accurate than BINSEG and LOPART (when using a supervised penalty learned from GFPOP). 
These benefits indicate that FLOPART should be used over LOPART when a user requires a peak detection model that contains train set labels. 
In the future we will be interested to generalize the ideas from FLOPART to changepoint models with constraints other than the alternating up-down constraints that we considered in this paper.


\section*{Broader Impact}
Like any algorithm used to analyze biomedical data, our proposed algorithm, FLOPART, comes with different benefits and drawbacks. FLOPART greatly benefits both patients and biologists by quickly and correctly detecting abrupt changes in biomedical data. FLOPART will also allow biologists to label easily identifiable regions while predicting regions that aren't as easy to identify. As seen in our empirical analysis, FLOPART sometimes has false positives and negatives, so it could potentially falsely predict regions of importance to a biologist. 
This could cause a biologist to assume areas of a patient's biomedical data are significant and vice versa. 
A large number of false positives/negatives are caused by small biases in the data (small changes in the sequence of counts in the genomic time series), and most of them can be corrected by our method of tuning the penalty hyper-parameter (but for some data sets, important regions could be missed in reality regardless of the penalty). 

\bibliographystyle{abbrvnat}
\bibliography{refs}

\end{document}